\documentclass[sigconf]{acmart}
\usepackage{enumitem}
\AtBeginDocument{%
  \providecommand\BibTeX{{%
    \normalfont B\kern-0.5em{\scshape i\kern-0.25em b}\kern-0.8em\TeX}}}

\copyrightyear{2021}
\acmYear{2021}
\setcopyright{acmcopyright}
\acmConference[CIKM '21] {Proceedings of the 30th ACM International Conference on Information and Knowledge Management}{November 1--5, 2021}{Virtual Event, Australia.}
\acmBooktitle{Proceedings of the 30th ACM Int'l Conf. on Information and Knowledge Management (CIKM '21), November 1--5, 2021, Virtual Event, Australia}
\acmPrice{15.00}
\acmISBN{978-1-4503-8446-9/21/11}
\acmDOI{10.1145/3459637.3482266}

\usepackage{balance}

\begin{document}
\fancyhead{}
\title{Inductive Matrix Completion Using Graph Autoencoder}



\author{Wei Shen}
\email{shenwei0917@126.com}
\authornote{The first two authors contribute equally to this work.}
\affiliation{%
  \institution{Nanjing University}
}

\author{Chuheng Zhang}
\email{zhangchuheng123@live.com}
\authornotemark[1]
\affiliation{%
  \institution{Tsinghua University}
}

\author{Yun Tian}
\email{tianyun@shanghaitech.edu.cn}
\affiliation{%
  \institution{ShanghaiTech University}
}

\author{Liang Zeng}
\email{zengl18@mails.tsinghua.edu.cn}
\affiliation{%
  \institution{Tsinghua University}
}

\author{Xiaonan He}
\email{xiaonan.cs@gmail.com}
\affiliation{%
  \institution{Beihang University}
}

\author{Wanchun Dou}
\email{douwc@nju.com}
\authornote{Corresponding author.}
\affiliation{%
  \institution{Nanjing University}
}

\author{Xiaolong Xu}
\email{xlxu@ieee.org}
\affiliation{%
  \institution{Nanjing University of Information Science and Technology}
}

\renewcommand{\shortauthors}{Trovato and Tobin, et al.}


\begin{abstract}
    Recently, the graph neural network (GNN) has shown great power in matrix completion by formulating a rating matrix as a bipartite graph and then predicting the link between the corresponding user and item nodes. The majority of GNN-based matrix completion methods are based on Graph Autoencoder (GAE), which considers the one-hot index as input, maps a user (or item) index to a learnable embedding, applies a GNN to learn the node-specific representations based on these learnable embeddings and finally aggregates the representations of the target users and its corresponding item nodes to predict missing links. However, without node content (i.e., side information) for training, the user (or item) specific representation can not be learned in the inductive setting, that is, a model trained on one group of users (or items) cannot adapt to new users (or items). To this end, we propose an inductive matrix completion method using GAE (IMC-GAE), which utilizes the GAE to learn both the user-specific (or item-specific) representation for personalized recommendation and local graph patterns for inductive matrix completion. Specifically, we design two informative node features and employ a layer-wise node dropout scheme in GAE to learn local graph patterns which can be generalized to unseen data. 
    The main contribution of our paper is the capability to efficiently learn local graph patterns in GAE, with good scalability and superior expressiveness compared to previous GNN-based matrix completion methods. Furthermore, extensive experiments demonstrate that our model achieves state-of-the-art performance on several matrix completion benchmarks. Our official code is publicly available\footnote{https://github.com/swtheing/IMC-GAE}.     
\end{abstract}



\maketitle
\section{Introduction}
Matrix completion (MC) \cite{mnih2008probabilistic, kalofolias2014matrix,dziugaite2015neural} is one of the most important problems in modern recommender systems, using past user-item interactions to predict future user ratings or purchases. Specially, given a partially observed user-item historical rating matrix whose entries represent the ratings of users with items, MC is to predict the missing entries (unobserved or future potential ratings) in the matrix based on the observed ones. The most common paradigm of MC is to factorize the rating matrix into the product of low-dimensional latent embeddings of rows (users) and columns (items), and then predict the missing entries based on these latent embeddings. Traditional matrix completion methods \cite{candes2009exact, dziugaite2015neural} have achieved great successes in the past. However, these methods mainly learn the latent user (or item) representation yet largely neglect an explicit encoding of the collaborative signal to reveal the behavioral similarity between users \cite{wang2019neural}. These signals are crucial for predicting the missing rating in the rating matrix, but hard to be exploited, since they are hidden in user-item interactions \cite{he2020lightgcn}. 

Recently, many works \cite{10.5555/3294996.3295127,berg2017graph,zhang2019inductive, you2020handling} have studied using a GNN to distill collaborative signals from the \emph{user-item interaction graph}. Specially, matrix completion is formulated as link prediction, where the rating matrix is formulated as a bipartite graph, with users (or items) as nodes and observed ratings/interactions as links. 
The goal of GNN-based matrix completion methods is to predict the potential or missing links connecting any pair of nodes in this graph. Graph Autoencoder (GAE) \cite{kipf2016variational} is a popular GNN-based link prediction method, where a GNN is first applied to the entire network to learn node-specific representations. Then the representations of the target nodes are aggregated to predict the target link. Many GNN-based matrix completion methods directly apply GAE to the rating graph to predict potential ratings such as GC-MC and NMTR \cite{berg2017graph, gao2019neural}. By exploiting the structure of the bipartite user-item graph, the node-specific representations learned by GAE, which represents user-specific preferences or item attributes, are more expressive than the patterns learned by the traditional matrix completion methods for personalized recommendation.

\begin{table}[!t]
    \centering
    \begin{tabular}{c|ccc}
      \toprule
         & GAE-based models & IGMC & IMC-GAE (ours)\\
      \midrule
      Specific & \checkmark & $\times$ & \checkmark\\
      Local & $\times$ & \checkmark & \checkmark \\
      Efficient & \checkmark & $\times$ & \checkmark\\
      Inductive & $\times$ & \checkmark & \checkmark\\
      \bottomrule
    \end{tabular}
    \vspace{1em}
    \caption{We compare the GNN-based matrix methods from different aspects: 1) whether they learn node-specific representations for personalized recommendation (denoted as Specific), 2) whether they learn local graph patterns (denoted as Local), (3) whether they are efficient matrix completion methods (denoted as Efficient), (4) whether they are inductive matrix completion methods (denoted as Inductive)}
    \vspace{-2.8em}
    \label{tab:comparison}
\end{table}

Despite its effectiveness, there remain two main challenges to apply GAE-based matrix completion to real recommender systems. The first challenge stems from a key observation from real-world scenarios: There are a large number of users or items in a real recommender system that have few historical ratings. This requires a model to predict potential ratings in a sparse rating matrix. However, GAE-based models usually fail in this situation since there are a few historical ratings in a sparse rating matrix for GAE-based models to train node-specific representations for personalized recommendation \cite{zhang2020revisiting}. The second challenge is applying the GAE-based models to real recommender systems for the large-scale recommendation. In real recommender systems, new users (or items) are emerging that are not exposed to the model during training. This requires that the model to be \emph{inductive}, i.e., the model trained on a group of users (or items) can adapt to new groups. However, previous GAE-based models are all transductive models so that the learned node representations cannot be generalized to users (or items) unseen during training \cite{zhang2019inductive}. 

The following question arises: Can we have a GAE-based model that can not only guarantee good performance on a sparse rating matrix but also enable inductive learning? In fact, using GAE to simultaneously satisfy the two requirements for matrix completion is a non-trivial challenge when high-quality user (or item) features are unavailable. The one-hot node indices (together with learnable node-specific embeddings) in the GAE-based model give a maximum capacity for learning distinct user preferences (or item attributes) from historical ratings. On the other side, learning distinct user preferences (or item attributes) in GAE also requires adequate rating samples from the rating matrix. Accordingly, without adequate rating samples in a sparse rating matrix, it is hard for GAE to obtain satisfactory performance. Moreover, for unseen nodes from a new rating matrix, GAE lacks the representations of them, and therefore cannot predict the potential ratings in a new rating matrix, which makes inductive learning impossible. To overcome these two challenges, \citet{zhang2019inductive} propose an inductive matrix completion based on GNN (IGMC). To predict a potential link (i.e., rating), it first extracts a 1-hop subgraph around the target link and then relabels the node w.r.t the distance to the target nodes. Finally, a GNN is applied to each subgraph to learn the local graph patterns that can be generalized to an unseen graph. By learning local graph patterns, IGMC has a better performance on the sparse rating matrix and enables inductive matrix completion. However, extracting subgraphs in both training and inference processes is time-consuming for the real recommendation. Moreover, the performance degradation on the dense rating matrix in IGMC also hinder us from applying it to real recommender systems. 

In this paper, we propose an inductive matrix completion method using GAE (IMC-GAE) that achieves efficient and inductive learning for matrix completion, and meanwhile obtain good performance on both sparse and dense rating matrices. As summarized in Table \ref{tab:comparison}, IMC-GAE combines the advantages of both the GAE-based models and IGMC together, which uses GAE to learn both node-specific representation for personalized recommendation, and local graph patterns for inductive matrix completion. Specially, we incorporate two informative node features into IMC-GAE to represent two types of user-item interactions and design a layer-wise node dropout scheme in GAE to learn local graph patterns for inductive matrix completion. 

In summary, this work makes the following main contributions:
\begin{itemize}[leftmargin=*]
    \item (Sec. \ref{sub_sec_1}) To better understand local graph patterns, we conduct a quantitative analysis on five real datasets. Based on this quantitative analysis, we have multiple observations that reveal the properties of local graph patterns in matrix completion. It motivates us to design our model, IMC-GAE. 
    \item (Sec. \ref{sub_sec_2}) We design two informative features, the identical feature and the role-aware feature, for the model to learn the expressive graph patterns. Moreover, these graph patterns can be easily generalized to unseen graphs.
    \item (Sec. \ref{sub_sec_3}) We design a layer-wise node dropout schema that drops out more nodes in the higher layers. With the layer-wise node dropout, link representation in our model contains more node information in a 1-hop local graph around the target link. Accordingly, our model is able to learn local graph patterns associated with the target link, which enhances the capability of the inductive learning of our model.
    \item (Sec. \ref{sub_sec_4}) To illustrate the effectiveness of the proposed IMC-GAE, we conduct empirical studies on five benchmark datasets. Extensive results demonstrate the state-of-the-art performance of IMC-GAE and its effectiveness in learning both local graph patterns and node-specific representations.
\end{itemize}

\begin{figure*}[!tp]
    \centering\includegraphics[width=\textwidth]{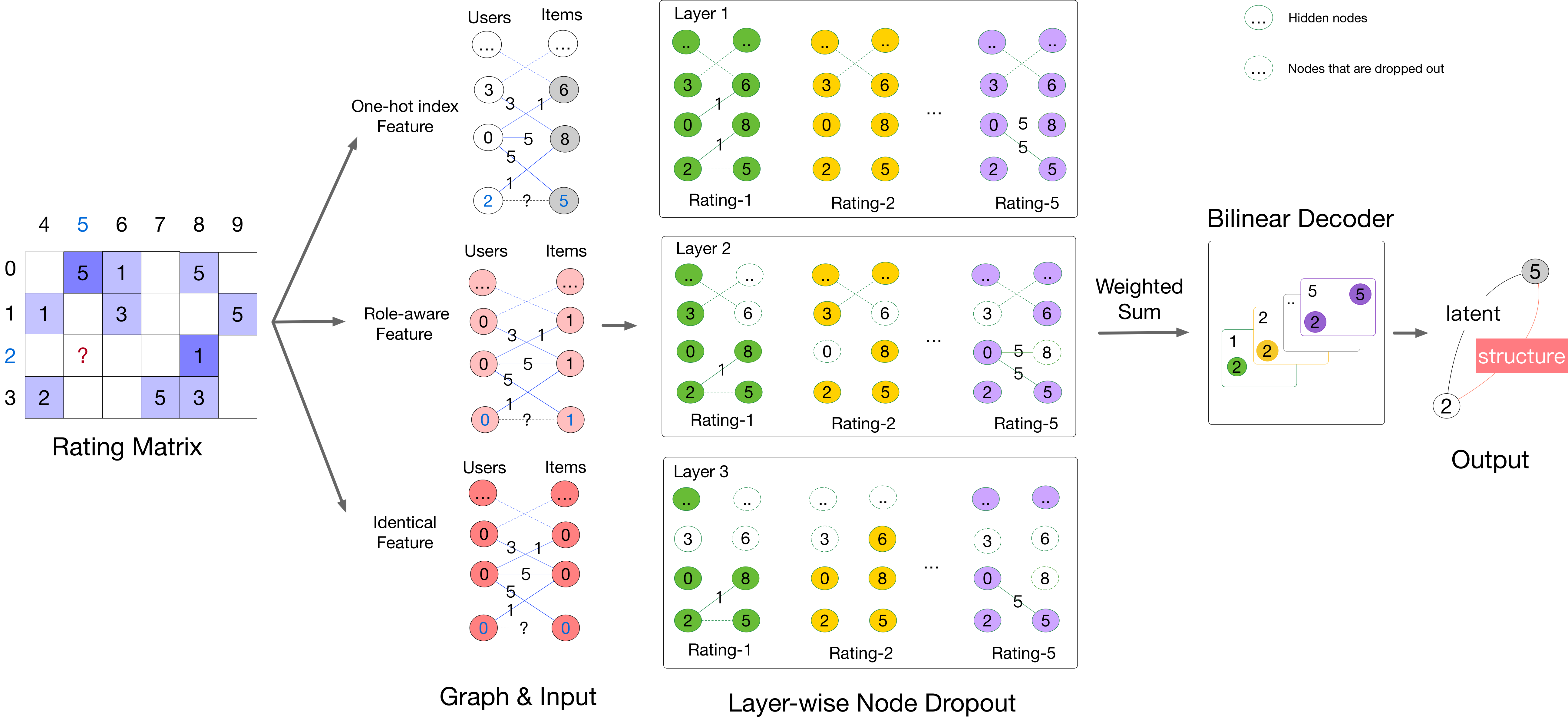}
    \caption{Model Overview. The rating matrix is formulated as a bipartite user-item graph, in which the nodes represent users (or items) and the links represent the corresponding ratings. 
    In addition, the input features of each node in this graph consist of the identical feature, the role-aware feature, and the one-hot index feature. 
    In addition, the encoder of our model has multiple layers (e.g., Layer 1) with multiple rating-subgraph (e.g., Rating 1). 
    As stacking more layers, the node dropout probability increases, which is referred to as layer-wise node dropout. 
    The model aggregated the latent embedding which is learned by one-hot index feature and structure embedding of a node which is learned by role-aware feature and identical feature in all layers by the weighted sum operator. 
    At last, we reconstruct the links by a bilinear decoder. In this way, the output of our model contains the information of both latent link representation and structure representation}
    \label{fig:Overview}
\end{figure*}

\vspace{-8pt}
\section{Related Works}
 In this section, we will briefly review existing works on GAE-based matrix completion methods and inductive matrix completion methods based on GNN, which are most relevant with this work. Here, we highlight their differences to IMC-GAE, and illustrate how we combine the advantages of them to build a more effective model for real recommendation.

\subsection{GAE-based matrix completion}
The majority of GNN-based matrix completion methods is based on Graph Autoencoder (GAE) \cite{kipf2016variational},  which applies a GNN to the entire network to learn a representation for each node. The representations of the user and item nodes are aggregated to predict potential ratings. For example, \citet{10.5555/3294996.3295127} propose a multi-graph CNN model to extract user and item latent features from their nearest-neighbor networks. \citet{berg2017graph} propose graph convolutional matrix completion (GC-MC) which uses one-hot encoding of node IDs as initial node features, learns specific node representations by applying a GNN-encoder to the bipartite user-item graph, and reconstructs the rating links by a GNN-decoder. To the best of our knowledge, our method is the first inductive GAE-based matrix completion method that achieves a good performance in both sparse and dense rating matrices.    
\subsection{Inductive GNN-based matrix completion methods}
There are mainly two types of GNN-based matrix completion methods that are applicable to inductive settings. One is attempting to handle inductive matrix completion without using node content, such as IGMC \cite{zhang2019inductive}. IGMC first extracts enclosing subgraphs around target links, then relabels the nodes in subgraphs according to their distances to the source and target nodes, and finally applies a GNN to each subgraph to learn a link representation for link prediction. IGMC applies GNN to those enclosing subgraphs to learn local graph patterns, which can easily generalize to the users (or items) unseen during training. Moreover, local graph patterns help IGMC obtain a better performance than the GAE-based models on the sparse rating matrices. However, applying IGMC to real recommender systems yields two crucial challenges. First of all,  IGMC replaces nodes' one-hot index embedding with local structure features, which does not capture diverse user preferences and item attributes for personalized recommendation. Second, IGMC extracts subgraphs around target links during both the training and inference process, which is time-consuming for large-scale recommendation. In contrast, IMC-GAE maintains the ability to give a node-specific representation, which is important in personalized recommendation for the users with historical ratings. In addition, instead of extracting subgraphs and relabeling each node, we incorporate two informative features into the input features of each node and design a layer-wise node dropout scheme in IMC-GAE to help the GAE to learn local graph patterns. By using GAE to learn local graph patterns, the inference process of IMC-GAE becomes efficient and inductive.

Another previous inductive GNN-based matrix completion methods are content-based models; such as PinSage \cite{ying2018graph}, which uses node content as initial node features. Although being inductive and successful in real recommender systems, content-based models rely heavily on the rich content of each node, which is not easily accessible in most real recommender systems. In comparison, our model is inductive and does not rely on any node content.
\section{Method}
As aforementioned, matrix completion has been formulated as the link prediction problem on a bipartite user-item graph in recent GNN-based matrix completion methods. 
Specially, we consider a matrix completion that deals with a rating matrix $M$ of shape $N_u \times N_v$, where $N_u$ is the number of users and $N_v$ is the number of items. Some entries in this matrix exist and other entries are missing. Existing entry $M_{ij}$ is a historical rating from a user $i$ to an item $j$. 
The task of matrix completion is to predict the value of missing entries. 
GNN-based matrix completion views the matrix as a bipartite graph and predicts the missing links in this graph.
In this section, we first present some findings on multiple real-world datasets, which reveal the properties of local graph patterns in both sparse and dense rating matrices. 
Based on these observations, we then elaborate on how the proposed learning algorithm, IMC-GAE, integrates the GAE-based model and IGMC to obtain a more effective model for real recommender systems. Then, we show an overview of IMC-GAE in Figure \ref{fig:Overview}. Specially, IMC-GAE is a GAE-based model consisting of three major components:  1) embedding layer whose input features consist of the one-hot index of nodes, the identical feature and the role-aware feature, 2) relational GCN encoder, 3) a bilinear decoder that combines the representations of target nodes to reconstruct links representation. 
\begin{table}
  \caption{Quantitative Analysis on multiple datasets.}
  \label{tab:QAD}
  \begin{tabular}{lcccccc}
    \toprule
    \textbf{Dataset} & \textbf{density} &\textbf{AUR}&\textbf{AIR}&\textbf{MCR}&\textbf{SCF} \\
    \midrule
    YahooMusic & < 0.0001 & 0.1915 & 0.0745 & 0.3585 & 0.4713\\
    Flixster & 0.0029 & 0.4705 & 0.1289 & 0.4362 & 0.5008\\
    Douban & 0.0152 & 0.3672 & 0.5033 & 0.4537 &  0.4735\\
    ML-1M & 0.0447 & 0.3771 & 0.4812 & 0.4151 &  0.5659\\
    ML-100K & 0.0630 & 0.3826 & 0.4177 & 0.3815 & 0.5006\\
  \bottomrule
\end{tabular}
\end{table}
\subsection{Understanding local graph patterns}
\label{sub_sec_1}
In the previous works, some handcrafted heuristics in a local graph around the target link (i.e., \textbf{local graph patterns}) are designed for link prediction on graphs \cite{liben2007link}. IGMC first adopts the labeling trick in GNN-based matrix completion that automatically learns suitable graph patterns from the local graph. 
These local graph patterns can be easily generalized to new local graphs or unseen links. 
To develop a better understanding of local graph patterns in matrix completion, we do a quantitative data exploration on five real-world datasets, the density of which ranges from less than $0.0001$ to $0.063$. In particular, we examine the Pearson's correlation coefficient (PCC) between the true ratings and four heuristic scores \cite{liben2007link, zeng2021graph}: average user rating (AUR), average item rating (AIR), most common rating between source nodes and target nodes (MCR) and a simple collaborative signal (SCF) in five datasets. Specially, we find a user node that has the most common neighbors with the source node as \emph{guider} in SCF. The link prediction in SCF is based on the rating that \emph{guider} rates the target item node. From Table \ref{tab:QAD}, we can extract multiple findings,
\begin{itemize}[leftmargin=*]
    \item The PCCs between the true ratings and four heuristic scores in five datasets are all positive, which indicates that the true ratings are correlated with these four heuristic scores in each dataset. Furthermore, it suggests that local graph patterns are effective to predict the missing ratings in matrix completion.
    \item The PCCs between the true ratings and four heuristic scores are all smaller than $6.0$, which indicates that a single local graph pattern is not enough to predict the missing ratings. It suggests that the model needs to learn more complex local graph patterns from rating matrix or specific node representations for personalized recommendation to obtain better performance.
    \item Among the four heuristic scores, AUR and AIR are simple statistics that only depend on one type of user-item interaction (i.e., the interactions with the target user or the interactions with the target item), while MCR is a statistic depending on these two types of interactions. We find that the performance of MCR is more stable across different datasets than that of AUR (or AIR) . It suggests that MCR is effective in both sparse and dense rating matrices. Moreover, stable local graph patterns like MCR are effective across different datasets, which makes inductive matrix completion possible. Furthermore, SCF considers all the interactions with the target nodes and their neighbors within 1-hop, which outperforms MCR on all datasets. It suggests that local graph patterns which consider more user-item interactions may be more powerful.  
\end{itemize}

\subsection{Input features}
\label{sub_sec_2}
Motivated by our earlier findings, we now introduce two input features, identical feature and role-aware feature for the GAE-based model to learn local graph patterns. The identical feature is an identical index, which helps GNN model aggregate one-hop user-item interactions (user-to-item interaction or item-to-user interaction) by message passing function. It aims to represent some simple local heuristics scores such as AIR or AUR, which have been demonstrated to be effective to predict potential ratings in the above quantitative analysis. 
To model two-hop user-item interactions, we design the second structure feature, the role-aware feature, using two extra indexes to distinguish user and item in the input space of the model. It helps the model distinguish user nodes with item nodes, and therefore distinguish the interactions from user to item with the interactions from item to user. Furthermore, after the user-item interactions around the target link are aggregated by the message passing function, the model can distinguish the user-item interactions from the 1-hop neighbors with the user-item interactions from 2-hop neighbors. By distinguishing these two types of user-item interactions, the model is capable of learning more complicated and powerful local graph patterns such as the aforementioned MCR or SCF. 

Furthermore, the model needs more expressive patterns for personalized recommendation. Accordingly, we incorporate the one-hot index into the input space of IMC-GAE, which is same as previous GAE-based models that learns specific node representations for personalized recommendation. Altogether, we adopt two informative features and one-hot index feature in IMC-GAE, which aims to help GAE learn structure link representation and latent link representation, respectively. The structure link representation represents local graph patterns around the target link, and the latent link representation represents the user-specific preference to the item.

\subsection{GNN encoder on heterogeneous graph}
In our paper, matrix completion is formulated as the link prediction problem on a heterogeneous graph. In the heterogeneous graph, rating edges of the same type are collected into a rating subgraph (e.g., if the graph consists of four types of ratings, there are four rating subgraphs). Correspondingly, each rating subgraph contains a copy of all the nodes. Then IMC-GAE applies a node-level GNN encoder to these subgraphs that learn a distinct representation for each node in each subgraph. There are three components in our GNN encoder: 1) embedding layer, 2) message passing layer, and 3) accumulation layer. 

\subsubsection{Embedding layer.} 
In each rating subgraph, the representation of each node consists of three different embeddings (identical node embedding $u_t$, role-aware embedding $r_t$, and rating embedding $l_t$). We assume that there are $T$ rating types in the rating matrix so that we have $T$ rating subgraphs in our model. With three different embeddings in each rating subgraphs, each node has $3 \times T$ embeddings in IMC-GAE. 
In order to reduce the number of parameters while allowing for more robust pattern learning, we use the same identical node embedding and role-aware embedding in each rating subgraph. 
Therefore, there are $T + 2$ embeddings to represent a node in $T$ rating subgraphs. Moreover, we concentrate (denoted by $Concat(\dot)$) these three embeddings (denoted by $U_t$, $R_t$, $L_t$) in embedding layer, which is the output of the embedding layer,
\begin{equation}
\label{eq:ini}
    x_t^0[i] = Concat(u_t[i], r_t[i], l_t[i]),
\end{equation}
where $x_t^0[i]$ denotes node $i$’s embedding vector in $t$-th rating subgraph. The node embedding vectors are the input of message passing layer.

\subsubsection{Message passing layer.} In IMC-GAE, we adopt a traditional GCN message passing layer to do local graph convolution, which has the following form:
\begin{equation}
    x_t^{l+1}[i] = \sum_{j \in \mathcal{N}_t(i)} \frac{1}{\sqrt{|\mathcal{N}_t(i)|\cdot |\mathcal{N}_t(j)|}}x_t^l[j]
\end{equation}
where $x_t^{l+1}[i]$ denotes node $i$’s feature vector at layer $l+1$ in the $t$-th rating subgraph. 
In addition, we chose symmetric normalization as the degree normalization factor in our message passing layer, where the $|\mathcal{N}_t(i)|$ represents the number of neighbors of node $i$ in the $t$-th rating subgraph. 

\subsubsection{Accumulation layer.} 
In each $t$-th rating subgraph, we stack $L$ message passage layer with ReLU activations \cite{agarap2018deep} between two layers. Following \cite{he2020lightgcn}, node $i$'s feature vectors from different layers are weighted sum as its final representation $h_t[i]$ in the $t$-th rating subgraph,
\begin{equation}
    h_t[i] = \sum_{0 \leq l \leq L}\frac{1}{l+1}x_t^1[i] 
\end{equation}
Then we accumulate all node $i$'s final representation $h_t[i]$ from all $T$ rating subgraphs into a single vector representation by sum operator,
\begin{equation}
    h[i] = \sum_{t\in T} h_t[i]
\end{equation}
To obtain the final representation of user or item node, we transform the intermediate output $h[i]$ by a linear operator,
\begin{equation}
    n[i] = tanh(Wh[i])
\end{equation}
The parameter matrix $W$ of user nodes is the same as that of item nodes, which because the model is trained without side information of the nodes.
\subsection{Bilinear decoder}
In IMC-GAE, following \cite{berg2017graph}, we use a bilinear decoder to reconstruct links in the user-item graph and treat each rating level as a separate class. Given the final representation $n[i]$ of user $i$ and $n[j]$ of item $j$, we use billinear operator to produce the final link representation $e_t[i, j]$ in the $t$-th rating subgraph,
\begin{equation}
    e_t[i, j] = n[i]^TW_tn[j],
\end{equation}
where $W_t$ is a learnable parameter matrix. Thus, we can estimate the final rating score as,
\begin{equation}
    r[i, j] = \sum_{t \in T} t S_t(\mathbf{e(i, j)}),
\end{equation}
where $\mathbf{e(i, j)}$ is the vector that concentrate the final link representations of user $i$ and item $j$ on all $T$ rating subgraph, and the $S_t$ is the softmax probability on $t$-th dimension of $\mathbf{e(i, j)}$ vector. 
\begin{figure}[tp]
    \centering\includegraphics[width=3.0in]{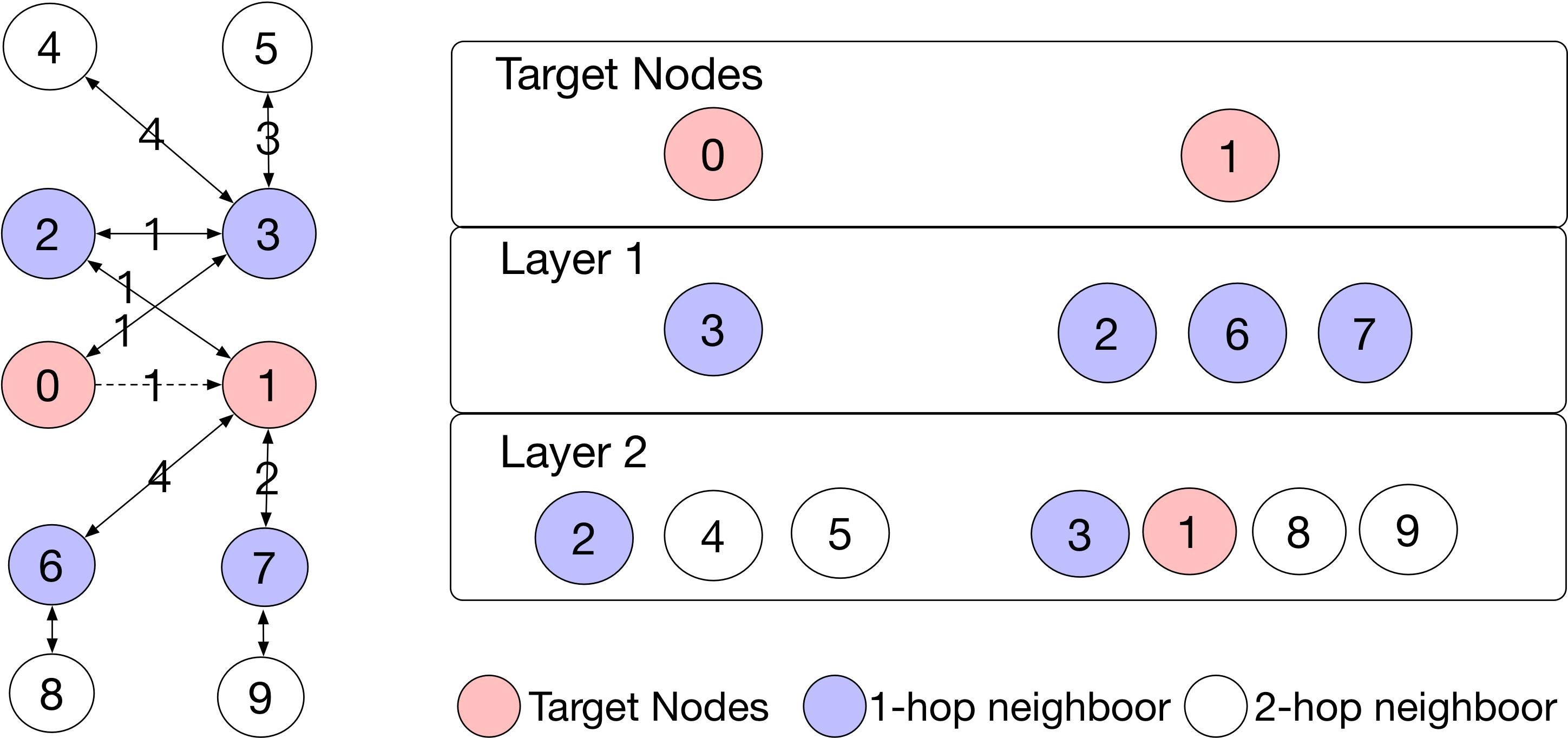}
    \caption{Layer-wise Node Dropout. In this subgraph extracted in ML-100k, red nodes indicates target nodes; blue nodes indicates the 1-hop neighbors of target nodes; white nodes indicates the 2-hop neighbors of target nodes.}
    \label{fig:NodeDropout}
\end{figure}
\subsection{Layer-wise node dropout}
\label{sub_sec_3}
The layer-wise node dropout is inspired by node dropout from \cite{berg2017graph}, aiming to help model grasp patterns in local graphs which can be better generalized to unobserved ratings.
In previous works, GAE-based models always adopt a node dropout scheme which randomly drops out all outgoing messages of a particular node with a probability $p$. However, our method adopts different node dropout probabilities in different layers, which we call it layer-wise node dropout. Specially, layer-wise node dropout is $p_l = p_0 - l \theta$,
where $p_l$ is the node dropout probability in the $l$-th layser, $p_0$ is the initial node dropout probability, and $\theta$ is the hyperparameter. 

In our paper, layer-wise node dropout facilitates node representation learning due to the following two reasons. 
The first reason is the same as \cite{berg2017graph}, which we adopt to overcome the over-smoothing problem in GNN representation and improve the generalization ability of our model. 
The second reason is to help the model learn local graph patterns which consider more user-item interactions in a 1-hop subgraph around the target link. As shown in Figure \ref{fig:NodeDropout}, the target nodes are node $0$ and node $1$. In the previous GAE-based models, the representations of node $0$ and node $1$ in the $2$-th layer aggregate too much node information beyond the $1$-hop subgraph around them (e.g., node $4$, node $5$, node $8$ and node $9$ in the example), which prevents the model learning graph patterns from the user-item interactions around target nodes. 
\begin{figure*}[tp]
    \centering\includegraphics[width=5.5in]{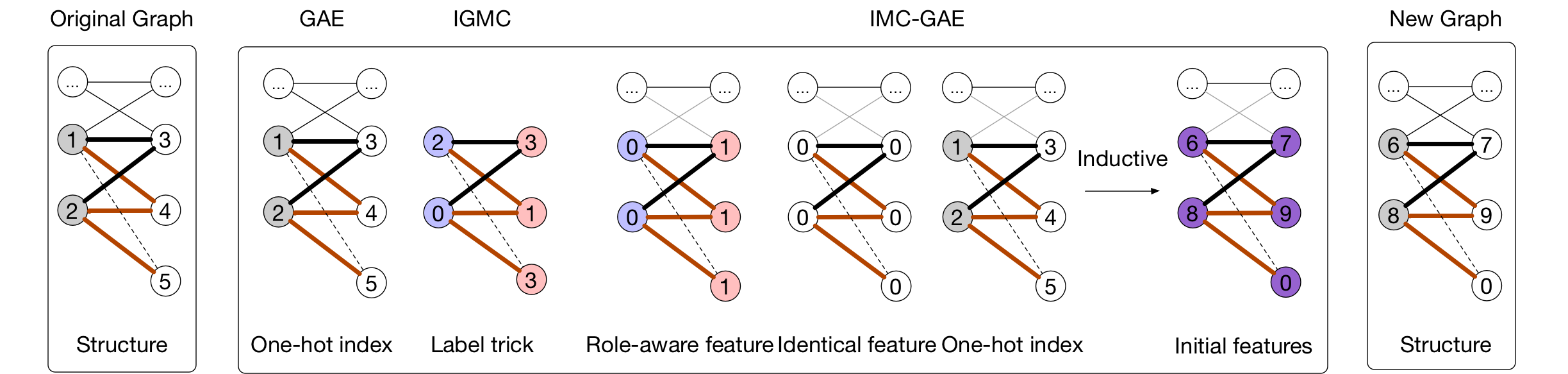}
    \caption{We compare local graph patterns learning in IMC-GAE with that in GAE and IGMC in two cases. In the first case, the model is to infer the link between node $1$ and node $5$ in original graph. In the second case, the model is to infer the link between node $6$ and node $0$ in a new graph.}
    \label{fig:Discuss}
\end{figure*}
\subsection{Model training}
\subsubsection{Loss function.} We minimize the cross entropy loss (denoted by $CE$) between the predictions and the ground truth ratings,
\begin{equation}
    \mathcal{L} = \frac{1}{|{(i, j)|\Omega_{i, j} = 1}|}\sum_{(i, j):\Omega_{i, j} = 1} CE(r[i, j], \hat{r}[i, j]),
\end{equation}
where we use $r[i, j]$ and $\hat{r}[i, j]$ to denote the true rating and the predicted rating of $(i, j)$, respectively, and the 0/1 matrix $\Omega$ serves as a mask for unobserved ratings in rating matrix $M$. 

\subsubsection{Node representation regularization.} 
It is inspired by adjacent rating regularization in \cite{zhang2019inductive}. 
Since each two rating types are comparable in matrix completion (e.g., ratings 5 is bigger than ratings 4 and ratings 4 is bigger than 3), we need to consider the magnitude of ratings. Accordingly,  we propose node representation regularization to encourages the representation of each node in rating subgraph that adjacent to each other to have similar representations. 
Specially, we assume that the representation of the $i$-th node in the $t$-th rating subgraph is $h_t[i]$, where $0 \leq t \leq T$. 
Then, the NRR regularizer is,
\begin{equation}
    \mathcal{L}_{NRR} = -\sum_{0 \leq t < T}\sum_{0 \leq i \leq N} Cos(h_t[i], h_{t+1}[i]),
\end{equation}
where $Cos$ is cosine similarity between two vectors, and
$N$ is the total number of users and items in the matrix. Finally, we combine these two loss functions to the final loss function,
\begin{equation}
\label{eq:loss}
    \mathcal{L}_{f} = \mathcal{L} + \lambda \mathcal{L}_{NRR},
\end{equation}
where $\lambda$ is a hyperparameter that trade-off two losses.

\subsection{Inductive learning}
In IMC-GAE, the inductive link representation for unseen nodes has two parts, inductive structure representation which is learned from the identical feature and the role-aware feature, and inductive latent representation which is learned from one-hot index of the node. 
For the inductive structure representation, we just leverage message passing, propagating learned structure representation from neighbors to target nodes. 
For the inductive latent representation, we also first accumulates the latent presentation of the neighbors of the target nodes. However, there may exist some unseen nodes during training in their neighbors, which lacks the latent representation. 
In our method, we use the average latent representation of the other nodes to represent the unseen nodes in each rating subgraph,
\vspace{-5pt}
\begin{equation}
    l_t[i] = \sum_{ j \in {\mathcal{I}}} \frac{1}{|{\mathcal{I}}|} l_t[j],
\end{equation}
where $l_t[i]$ is the initial latent representation of the $i$-th node in $t$-th rating subgraph (in equation \ref{eq:ini}) and $\mathcal{I}$ is a set of nodes which we have seen during training. It is a simple but effective method, which is demonstrated in the following experiments.

\section{Discussing local graph patterns}
To shed more light on local graph patterns learning in GNN-based models, we provide a comparison with GAE-based matrix completion, IGMC, and IMC-GAE through a typical example in Figure \ref{fig:Discuss}. Here, we assume the ratings in our example are within \{1, -1\} (like, denoted by  bold black line, and dislike, denoted by bold coffee line). The solid lines are observed ratings for training and dash lines are test ratings. In a train case, user $1$ and $2$ all like item $3$, while they all dislike item $4$. It indicates that user $1$ may have a similar taste with user $2$, which is a common local graph pattern in matrix completion. Furthermore, since user $2$ dislikes item $5$, we inference that user $1$ may dislike item $5$ based on the "similar taste" pattern.

When trained with the existing rating between user $2$ and item $4$, IGMC first extracts the 1-hop local graph around user $2$ and item $4$, and relabels user $1$, user $2$, item $3$, item $4$ and item $5$ as index $2$, $0$, $3$, $1$ and $3$, respectively. Finally, the model applies a GNN to the local graph, where the new node labels are the input features of the model. Without introducing the user-item interactions beyond the 1-hop local graph around target nodes, the "similar taste" pattern is easily learned by the model. However, previous GAE-based models apply the GNN to the entire graph for learning graph patterns. Accordingly, when trained with the existing rating between user $2$ and item $4$, the representations of user $2$ and item $4$ are aggregated with many node embeddings beyond 1-hop local graph around user $2$ and item $4$, which makes the model hardly focus on the interactions in this local graph, and fails to learn "similar taste" pattern. To solve this problem, IMC-GAE designs the layer-wise node dropout scheme for the GAE-based model to avoid aggregating too many embeddings of the nodes beyond the 1-hop local graph into the target nodes representation. Although the target node representations are still aggregated a few node representations beyond the local graph, the model is capable of learning the "similar taste" pattern between user $1$ and $2$. Furthermore, if given a new graph in Figure \ref{fig:Discuss} which has the same graph structure as the original graph, previous GAE-based models need to be retrained to inference the missing rating between user $8$ and item $9$. However, 
with labeling trick in IGMC and inductive structure representation in IMC-GAE, the models learn the "similar taste" pattern into structural link representation, which can be generalized to the new graph.

Despite the effectiveness of local graph patterns learning in IGMC, the labeling trick introduces extra computational complexity. The reason is that for every rating $(u_j, i_k)$ to predict, IGMC needs to relabel the graph according to $(u_j, i_k)$. The same node $u_j$ will be labeled differently depending on the target link and will be given a different node representation by the GNN when it appears in different links’ labeled graphs. This is different from previous GAE-based models and IMC-GAE, where we do not relabel the graph and each node only has a single embedding vector. For a graph with $n$ nodes and $m$ ratings to predict, the GAE-based model needs to apply the GNN $\mathcal{O}(n)$ times to compute an embedding for each node, while IGMC needs to apply the GNN $\mathcal{O}(m)$ times for all ratings. When $m \gg n$, IGMC has worse time complexity than GAE-based models, which is not suitable for real recommendation. 



\section{Experiments}
\label{sub_sec_4}
We perform experiments on five datasets to evaluate our proposed method. We aim to answer the following research questions:
\begin{itemize}[leftmargin=*]
    \item \textbf{RQ1:} How does IMC-GAE perform compared with state-of-the-art matrix completion methods when facing both sparse and dense rating matrices?
    \item \textbf{RQ2:} How does the different hyper-parameter settings (e.g., depth
of layer, weighted layer combination, and node representation regularization (NRR)) affect IMC-GAE?
    \item \textbf{RQ3:} How does the local graph patterns learning in IMC-GAE benefit from two informative features and layer-wise node dropout scheme respectively?
    \item \textbf{RQ4:} How does the IMC-GAE perform on few-shot or even unseen users (or items) as compared with GAE-based models and IGMC?
\end{itemize}

\subsection{Datasets description}
To evaluate the effectiveness of IMC-GAE, we conduct experiments on five common matrix completion datasets, Flixster \cite{jamali2010matrix}, Douban datasets \cite{ma2011recommender}, YahooMusic \cite{dror2012yahoo}, MovieLens-100K \cite{miller2003movielens} and MovieLens-1M \cite{miller2003movielens}, which are publicly accessible and vary in terms of domain, size, and sparsity. Moreover, Flixster, Douban, and YahooMusic are preprocessed subsets of the original datasets provided by \cite{monti2017geometric}. These datasets contain sub rating matrix of only 3000 users and 3000 items, which we consider as sparse rating matrices in real recommendation. The MovieLens-100K and MovieLens-1M are widely used datasets for evaluating many recommender tasks, which we consider as dense rating matrices in real recommendation. For ML-100k, we train and evaluate on canonical u1.base/u1.test train/test split. For ML-1M, we randomly split into 90\% and 10\% train/test sets. For the Flixster, Douban, and YahooMusic, we use the splits provided by \cite{monti2017geometric}. 

\begin{table}
  \caption{RMSE of different algorithms on Flixster, Douban and YahooMusic.}
  \label{tab:rmse_spare}
  \begin{tabular}{lccccc}
    \toprule
    \textbf{Model} &\textbf{Flixster}&\textbf{Douban}&\textbf{YahooMusic}\\
    \midrule
    IGC-MC  & 0.999 & 0.990 & 21.3\\
    F-EAE & 0.908 & 0.738 & 20.0\\
    PinSage & 0.954 & 0.739 & 22.9\\
    IGMC &\textbf{0.872} & \textbf{0.721} & 19.1\\
    \midrule
    GRALS  & 1.245 & 0.883 & 38.0\\
    sRGCNN & 0.926 & 0.801 & 22.4\\
    GC-MC & 0.917 & 0.734 & 20.5\\
    IMC-GAE (ours) & 0.884 & \textbf{0.721} & \textbf{18.7}\\
  \bottomrule
\end{tabular}
\end{table}

\begin{table}
  \caption{RMSE test results on MovieLens-100K (left) and MovieLens-1M (right).}
  \label{tab:rmse_dense}
  \begin{tabular}{lc|lc}
    \toprule
    \textbf{Model}&\textbf{ML-100K}&\textbf{Model}&\textbf{ML-1M}\\
    \midrule
    F-EAE & 0.920 & F-EAE & 0.860\\
    PinSage & 0.951 & PinSage & 0.906\\
    IGMC & 0.905 & IGMC & 0.857\\
    \midrule
    MC & 0.973 & PMF & 0.883\\
    IMC & 1.653 & I-RBM & 0.854\\
    GMC & 0.996 & NNMF & 0.843\\
    GRALS   & 0.945 & I-AutoRec & 0.831\\
    sRGCNN & 0.929 & CF-NADE & \textbf{0.829}\\
    GC-MC & 0.905 & GC-MC &  0.832\\
    NMTR & 0.911 & NMTR & 0.834 \\
    IMC-GAE (ours) & \textbf{0.897} & IMC-GAE(ours) & \textbf{0.829}\\
  \bottomrule
\end{tabular}
\end{table}
\subsection{Experimental Settings}
\subsubsection{Baselines.} 
To demonstrate the effectiveness, we compare our proposed IMC-GAE with the following methods:    
\begin{itemize}[leftmargin=*]
    \item \textbf{Traditional methods.} matrix completion (MC) \cite{candes2009exact}, inductive matrix completion (IMC) \cite{jain2013provable}, geometric matrix completion (GMC) \cite{kalofolias2014matrix}, PMF \cite{mnih2007probabilistic}, I-RBM \cite{salakhutdinov2007restricted}, NNMF \cite{dziugaite2015neural}, I-AutoRec \cite{sedhain2015autorec} and CF-NADE \cite{zheng2016neural} are traditional matrix competion methods, which use the user-item ratings (or interactions) only as the target value of their objective function.
    \item \textbf{GAE-based methods.} sRGCNN \cite{10.5555/3294996.3295127}, NMTR \cite{gao2019neural}, GC-MC \cite{berg2017graph} are GAE-based matrix completion methods, which use one-hot index as the initial feature of each node.
    \item \textbf{IGMC.} IGMC \cite{zhang2019inductive} is an inductive matrix completion method, which learns local graph patterns to generalize to new local graphs for inductive learning.
    \item \textbf{Content-based GNN methods.} Content-based matrix completion methods are inductive GNN-based methods adopting side information as initial features of each node, which includes PinSage \cite{ying2018graph} and IGC-MC \cite{berg2017graph}. PinSage is originally used to predict related pins and is adapted to predicting ratings here. IGC-MC is a content-based GC-MC method, which uses the content features instead of the one-hot encoding of node IDs as its input features.
    \item \textbf{Other GNN methods.} GRALS \cite{rao2015collaborative} is a graph regularized matrix completion algorithm and F-EAE \cite{hartford2018deep} uses exchangeable matrix layers to perform inductive matrix completion without using content.  
\end{itemize}
 In addition, given different datasets, we compare IMC-GAE with different baseline methods under RMSE in Table \ref{tab:rmse_spare}. The RMSE is a common evaluation metric in matrix completion \cite{zhang2019inductive, berg2017graph}. The baseline results are taken from \cite{zhang2019inductive}.  
 
\subsubsection{Hyperparameter Settings.} We implement our model based on DGL \cite{wang2019deep} and use the Adam optimizer. We apply a grid search for hyperparameters, the number of layers is searched in $\{1, 2, ... , 5\}$, the $\lambda$ in equation \ref{eq:loss} is searched in $\{4e^{-5}, 4e^{-4}, 4e^{-3}, 4e^{-2}\}$, the embedding size of each vector in embedding layer is chosen from $\{90, 120,..., 1800\}$ and the embedding size of each vector in bilinear decoder is searched in $\{30, 40, ... , 80\}$. Besides, the initial node dropout probability is tuned in $\{0.1, 0.2, 0.3\}$ and the decay ratio $\theta$ is tuned in $\{0.05, 0.1, 0.2\}$. 
All implementation codes can be found at \url{https://github.com/swtheing/IMC-GAE}.


\subsection{Performance comparison (RQ1)}
We start by comparing our proposed IMC-GAE with baselines on five benchmark datasets and then explore how the combination of the local graph patterns learning and specific node representations improves the performance in matrix completion.

For Flixster, Douban, and Yahoomusic, we compare our proposed model with GRALS, sRGCNN, GC-MC, F-EAE, PinSage, IGC-MC and IGMC. We show the result in Table \ref{tab:rmse_spare}. Our model achieves the smallest RMSEs on Douban and YahooMusic datasets, but slightly worse than IGMC on the Flixster dataset. Furthermore, as a GAE-based model, our method outperforms significantly all the GAE-based baselines (sRGCNN and GC-MC), which highlights the successful designs (two informative features, layer-wise dropout scheme) of our model.    

For ML-100k, we compare IMC-GAE with MC, IMC, as well as GRALS, sRGCNN, GC-MC, F-EAE, PinSage, NMTR and IGMC. For ML-1M, besides the baselines GC-MC, F-EAE, PinSage, NMTR and IGMC, we further include PMF, I-RBM, NNMF, I-AutoRec, and CF-NADE. Our model achieves the smallest RMSEs on these datasets without using any content, significantly outperforming all the compared baselines, regardless of whether they are GAE-based models. 

Altogether, our model outperforms all GAE-based models on all datasets. It demonstrates that the local graph patterns learned in our model truly help model inference the missing ratings in both sparse and dense rating matrices. 


\subsection{Study of IMC-GAE (RQ2)}
As the GNN encoder plays a pivotal role in IMC-GAE, we investigate its impact on the performance. We start by exploring the influence of layer numbers. We then study how the weighted layer combination and NRR affect the performance. 
\begin{table}
  \caption{Effect of layer numbers in GNN encoder}
  \label{tab:Layers}
  \begin{tabular}{lccc}
    \toprule
     &\textbf{Douban}&\textbf{YahooMusic} & \textbf{ML-100k}\\
    \midrule
    IMC-GAE-1 & 0.728 & 18.803 & 0.900 \\
    IMC-GAE-2 & 0.725 & 18.793 & \textbf{0.897} \\
    IMC-GAE-3 & 0.722 & \textbf{18.702} & 0.897 \\
    IMC-GAE-4 & 0.723 & 20.343 & 0.901 \\
    IMC-GAE-5 & \textbf{0.721} & 18.785 & 0.897 \\
  \bottomrule
\end{tabular}
\end{table}

\begin{table}
  \caption{Effect of weighted layer combination and NRR}
  \label{tab:Ablation}
  \begin{tabular}{lccc}
    \toprule
     &\textbf{Douban}&\textbf{YahooMusic}& \textbf{ML-100k}\\
    \midrule
    IMC-GAE(Original) & \textbf{0.721}& \textbf{18.7}&  \textbf{0.897}\\
    IMC-GAE(no NRR) & 0.722& 18.8&  0.900\\
    IMC-GAE(Sum) & 0.727 & 19.2&  0.905\\
    IMC-GAE(Concat) & 0.723& 18.9&  0.903\\
  \bottomrule
\end{tabular}
\end{table}
\subsubsection{Effect of Layer Numbers.} To investigate whether IMC-GAE can benefit from multiple layers in the GNN encoder, we vary the model depth. In particular, we search the layer numbers in the range of $\{1, 2, 3, 4, 5\}$. Table \ref{tab:Layers} summarizes the experimental results, wherein IMC-GAE-$i$ indicates the model with $i$ embedding propagation layers, and similar notations for others. By analyzing Table 
\ref{tab:Layers}, we have the following observations:
\begin{itemize}[leftmargin=*]
    \item Increasing the depth of IMC-GAE substantially enhances the performance of the model. Clearly, IMC-GAE-2 achieves consistent improvement over IMC-GAE-1 across all the board, which considers the 1-hop neighbors only. We attribute the improvement to the effective modeling of local graph structure: structure features and layer-wise node dropout help model grasp effective patterns from local graph around target nodes.
    \item When further stacking propagation layer on the top of IMC-GAE-2, we find that IMC-GAE-3 leads to performance degradation on ML-100k, but performance improvement on Douban and YahooMusic. This might be caused by the deeper layer with a higher node dropout probability might introduce noise in latent link representation. More specifically, the deeper layers (e.g., the third layer in IMC-GAE-3) lose the original graph connectivity, which makes the model fail to learn the latent link representation from the neighbors. Moreover, the marginal improvements on the other two datasets verify that the local graph patterns beyond 1-hop neighbors still improve the performance of the model in the sparse rating matrix.   
\end{itemize}
\subsubsection{Effect of weighted layer combination and NRR} Different from prior works \cite{berg2017graph,zhang2019inductive}, we adopt the weighted sum operator for layer combination instead of sum or concentration operator and NRR to encourages node representation in rating subgraph that adjacent to each other to have similar parameter matrice. Table \ref{tab:Ablation} shows the result of the ablation experiments. From the ablation experiments, we have the following observations. First, the weighted sum combination shows a performance improvement over the sum or concentration operator in both sparse and dense rating matrices. This might be because that sum or concentration operator does not assign lower importance to the node representations in deeper layers, which introduces more noise into the representation of the nodes. Second, we can see that disabling NRR results in the performance drop on all three datasets. It demonstrates that NRR is an effective way to regularize the model.
\begin{table}
  \caption{Ablation study on three datasets, where IMC-GAE-R indicates the IMC-GAE-R trained with only role-aware feature; IMC-GAE-I indicates IMC-GAE trained with only identical feature; IMC-GAE-OD indicates IMC-GAE trained with original node dropout scheme.}
  \label{tab:structure_cmp}
  \begin{tabular}{lccc}
    \toprule
    \textbf{Model}&\textbf{Douban}&\textbf{ML-100K}&\textbf{ML-1M}\\
    \midrule
    IMC-GAE & \textbf{0.721} & \textbf{0.897} & \textbf{0.829}\\
    IMC-GAE-R & 0.734 & 0.912 & 0.868\\
    IMC-GAE-I & 0.738 & 0.924 & 0.912\\
    IMC-GAE-OD & 0.727 & 0.905 & 0.834 \\
  \bottomrule
\end{tabular}
\end{table}
\begin{table}
  \caption{Inference time~(s) of IMC-GAE, IGMC and GCMC on Douban, MovieLens-100K, and MovieLens-1M.}
  \label{tab:infer}
  \begin{tabular}{lccc}
    \toprule
    \textbf{Model}&\textbf{Douban}&\textbf{ML-100K}&\textbf{ML-1M}\\
    \midrule
    IGMC & 9.255 & 33.041 & 122.042\\
    GC-MC & 0.011 & 0.011 &  0.025\\
    IMC-GAE(ours) & \textbf{0.060} & \textbf{0.0382} & \textbf{0.067}\\
  \bottomrule
\end{tabular}
\end{table}

\subsection{Study of the local graph patterns learning (RQ3)}
\subsubsection{Performance Comparison.} In this section, we attempt to understand how the identical feature, the role-aware feature, and the layer-wise node dropout scheme affect local graph patterns learning in IMC-GAE, and how local graph patterns learning affects the performance of IMC-GAE in both sparse and dense rating matrices. Towards this end, we compare the performance of original IMC-GAE with IMC-GAE trained with only identical feature, IMC-GAE trained with only role-aware feature, IMC-GAE with normal node dropout on the three datasets.
From the result shown in Table \ref{tab:structure_cmp}, we conclude the following findings:
\begin{itemize}[leftmargin=*]
    \item  With only role-aware feature or only identical feature for training, IMC-GAE obtains a competitive performance with IMC-GAE. It demonstrates that local graph patterns learning in IMC-GAE is effective in both sparse and dense rating matrices.
    \item The IMC-GAE outperforms the other three baselines in all datasets, which demonstrates that each design in IMC-GAE for local graph patterns learning (i.e., role-aware feature, identical feature, and layer-wise node dropout scheme) is essential, which helps GAE learn a series of effective local graph patterns. 
\end{itemize}
\subsubsection{Inference Time Comparison.} We compare the inference time of IMC-GAE with IGMC and GC-MC (a typical GAE-based model) on three datasets. Specially, we infer the 20\% samples in each dataset and conduct the experiment on GN7 on Tencent Cloud, which is equipped with 4*Tesla T4. We repeat this experiment five times and report the average inference time of each model in Table \ref{tab:infer}. The results show that the inference time of IMC-GAE is slightly longer than that of GAE but significantly shorter than that of IGMC.

\begin{figure}[tp]
    \centering\includegraphics[width=3.0in]{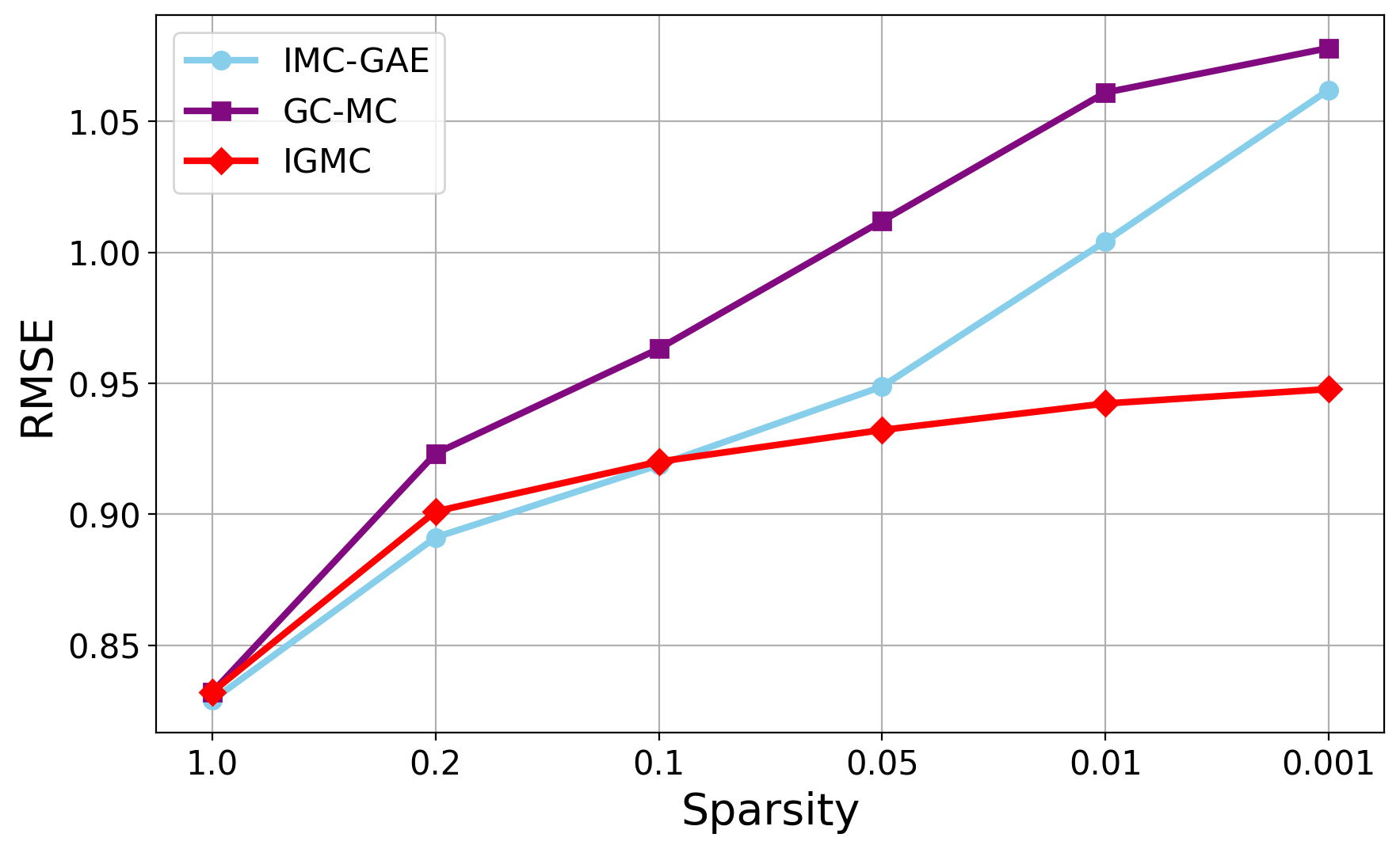}
    \caption{ML-1M results under different sparsity ratios.}
    \label{fig:sparse}
\end{figure}
\subsection{Study of IMC-GAE on sparse data (RQ4)}
To investigate the model performance on few-shot or even unseen users (or items), we test the model on dataset under different sparsity levels of the rating matrix \cite{berg2017graph, zhang2019inductive}. Here we construct several sparse datasets by using 100\%, 20\%, 10\%, 5\%, 1\% and 0.1\% training ratings in Movielens-1M, and then compare the test RMSEs of our method with GC-MC and IGMC, a typical GAE-based model and an inductive GNN model. As shown in Figure \ref{fig:sparse}, we have two observations, 
\begin{itemize}[leftmargin=*]
\item As the dataset becomes sparser, the performance of all the models suffer from a drop, but the drop rate of our model is much smaller compared with GC-MC. 
\item The performance of our model in the datasets with 100\%, 20\%, and 10\% training ratings is better than IGMC, but worse in other datasets.
\end{itemize}

From the observations, we find that the way we adopt GAE to learn local graph patterns that truly improves its inductive learning ability. However, the performance of IMC-GAE on three sparer datasets is worse than that of IGMC. It suggests that local graph patterns learned in IMC-GAE is not as good as those learned in IGMC which can be generalized to sparser datasets containing more few-shot or unseen users (or items). 

\section{Conclusion}
In this paper, we propose Inductive Matrix Completion using Graph Autoencoder (IMC-GAE), which uses GAE to learn both graph patterns for inductive matrix completion and specific node representations for personalized recommendation. Extensive experiments on real-world datasets demonstrate the rationality and effectiveness of the way IMC-GAE learns local graph patterns by GAE. This work represents an initial attempt to exploit local structural knowledge in GAE-based matrix completion, which is more suitable to be applied to real recommender systems.
\clearpage
\balance
\bibliographystyle{ACM-Reference-Format}
\bibliography{acmart}

\end{document}